\pdfoutput=1
\documentclass[11pt]{article}
\usepackage{acl}
\usepackage{times}
\usepackage{latexsym}
\usepackage[T1]{fontenc}
\usepackage[utf8]{inputenc}
\usepackage{microtype}
\usepackage{inconsolata}
\usepackage{hyperref}
\usepackage{graphicx}
\usepackage{array}
\usepackage{multirow}
\usepackage{amsmath}
\usepackage{relsize}
\title{Enabling and Analyzing How to Efficiently Extract Information from Hybrid Long Documents with LLMs}

\author{
    Chongjian Yue$^{1}$\thanks{~~Equal contribution and work done as interns at Microsoft Research Asia.}, Xinrun Xu$^{3}$\footnotemark[1], Xiaojun Ma$^{2, }$\thanks{~~Corresponding author}, Lun Du$^{2, }$\footnotemark[2],\\
    \textbf{Hengyu Liu}$^4$, \textbf{Zhiming Ding}$^{3}$, \textbf{Yanbing Jiang}$^{1}$, \textbf{Shi Han}$^{2}$, \textbf{Dongmei Zhang}$^{2}$\\
    $^{1}$School of Software \& Microelectronics, Peking University; $^{2}$Microsoft;\\
    $^{3}$Institute of Software Chinese Academy of Sciences; $^4$University of Technology Sydney\\
    \texttt{chongjian.yue@stu.pku.edu.cn, xuxinrun20@mails.ucas.ac.cn,} \\
    \texttt{\{xiaojunma, lun.du, shihan, dongmeiz\}@microsoft.com,} \\
    \texttt{hengyu.liu@uts.edu.au, zhiming@iscas.ac.cn, jyb@ss.pku.edu.cn}
}

\usepackage{listings}  
\usepackage{xcolor}  
\usepackage[T1]{fontenc}
\usepackage{mdframed} 

\definecolor{dkgreen}{rgb}{0,0.6,0}
\definecolor{gray}{rgb}{0.5,0.5,0.5}
\definecolor{light-gray}{gray}{0.95} 
\definecolor{mauve}{rgb}{0.58,0,0.82}
\definecolor{backcolour}{rgb}{0.95,0.95,0.92}

\newmdenv[linecolor=light-gray,
backgroundcolor=light-gray,
innerleftmargin=2.8pt,
innerbottommargin=-0.8pt,
leftmargin=0.0pt,
rightmargin=0.0pt,
skipbelow=-2.0pt,
frametitle={}]{codeframe}

\lstdefinestyle{prompts}{  
    commentstyle=\color{dkgreen},  
    keywordstyle=\color{magenta},  
    moredelim=**[is][\color{mauve}]{@}{@},  
    basicstyle=\fontsize{7}{9}\selectfont\ttfamily,  
    breakatwhitespace=false,  
    breaklines=true,  
    breakindent=0pt, 
    captionpos=b,  
    keepspaces=true,  
    numbers=none,  
    numbersep=5pt,  
    showspaces=false,  
    showstringspaces=false,  
    showtabs=false,  
    tabsize=2  
} 
\definecolor{mycolor}{rgb}{0.122, 0.435, 0.698}

\begin{document}
\maketitle
\begin{abstract}
Large Language Models (LLMs) demonstrate exceptional performance in textual understanding and tabular reasoning tasks.
However, their ability to comprehend and analyze hybrid text, containing textual and tabular data, remains unexplored.
Due to the hybrid text often appears in the form of hybrid long documents (HLDs), which is far exceed the token limit of LLMs. 
Consequently, we apply an naive split-recombination-based framework (SiReF) to enable LLMs process the HLDs and carry out experiments to analysis four important aspects of information extraction from HLDs.
Given the findings:
1) The effective way to select and summarize the useful part of a HLD.
2) An easy table serialization way is enough for LLMs to understand tables.
3) The naive SiReF has adaptability in many and complex scenarios.
4) The useful prompt engineering to enhance LLMs on HLDs.
To address the issue of dataset scarcity in HLDs and support the future work, we also propose the \textbf{Fi}nancial Reports \textbf{N}umerical \textbf{E}xtraction (\textbf{FINE}) dataset.
The dataset and code are publicly available in the attachments.
\end{abstract}

\begin{figure*}[ht]  
    \centering  
    \includegraphics[width=0.9\textwidth]{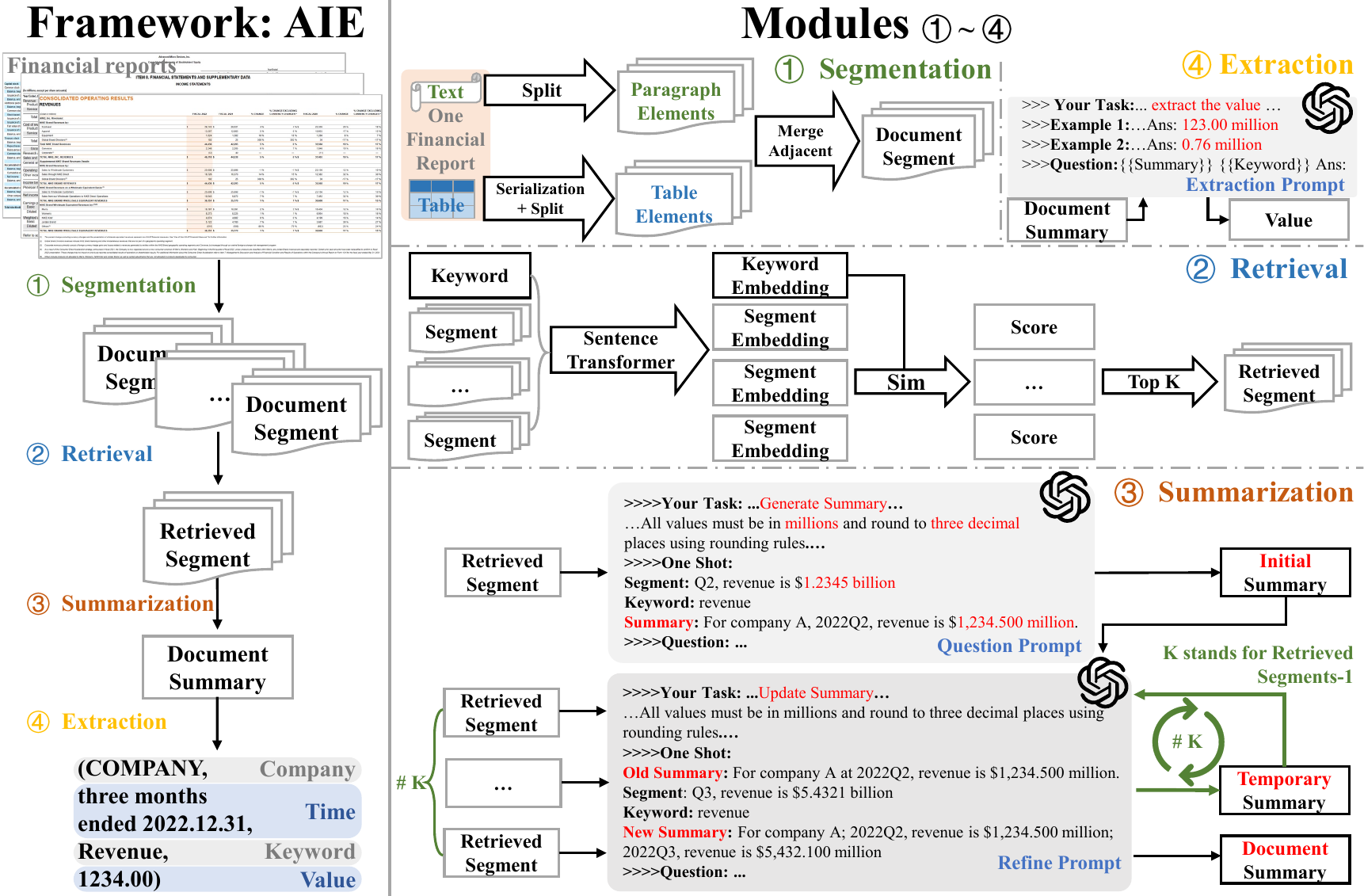}  
    \caption{The AIE framework illustrates the end-to-end IE process, consisting of four modules: Segmentation, dividing lengthy documents into short segments; Retrieval, selecting the most relevant segments related to the given keyword; Summarization, using LLMs to generate a concise summary of relevant information; and Extraction, extracting the keyword-corresponding value from the summary. This framework is exemplified using financial reports.}
    \label{fig:framework}  
    \vspace{-0.2cm}
\end{figure*}

\section{Introduction}
LLMs have exhibited remarkable capabilities in various natural language tasks, demonstrating their potential to comprehend and process intricate textual data~\cite{wei2023chainofthought,wang2023selfconsistency,zhou2022least,kojima2023large}. 
In addition to their success in textual data, the studies by~\cite{chen2023large,ye2023large} highlight the effectiveness of LLMs in handling tabular data. 
However, despite their proven proficiency in understanding and analyzing textual and tabular data individually, research exploring the capacity of LLMs to tackle hybrid documents, which combine these two types of data, remains relatively scarce.

Hybrid documents, adeptly blending textual and tabular content, are widely used across diverse fields and extensive in length typically.
Extracting relevant information from hybrid long documents (HLDs) based on user-provided keywords is a crucial upstream task that supports various applications, such as question-answering systems~\cite{DBLP:journals/csr/Gil23}, document classification~\cite{DBLP:journals/access/MustafaRASAAA23}, information retrieval~\cite{DBLP:journals/iwc/SilvaMO23}, and more.

Considering the excessive length of HLDs, LLMs can't directly process all the content in one HLD. 
If HLDs are subjected to simple truncation, substantial information loss will occur, resulting in compromised performance.
So we apply an easy split-recombination-based framework (SiReF), which splits the document into many segments and extracts information from the retrieved segments. 
Within the SiReF, the following challenges of HLDs are expected to be addressed effectively by LLMs:

\textbf{F1.} In HLDs, keyword-related information is distributed in many segments.
How we effectively select and summarize the useful segments? 
We carry out experiment from two dimensions: 
1) We compare two different summarization strategies: Refine and Map-Reduce. Refine shows a better accuracy, while Map-Reduce shows a better efficiency.
2) We investigate the effect of the number of retrieved segments based  of the similarity with keyword. 

\textbf{F2} Tables are commonly utilized in HLDs, but LLMs cannot directly interpret tabular data.
What would be an optimal serialization format for LLMs to comprehend this information more effectively?
We compare four different table serialization formats, and discovered that a simplified serialization format, devoid of much hierarchical information, is sufficient for LLMs.

\textbf{F3} For the adaptability, HLDs is a complex domain. How about the adapability of SiReF?
We carry out experiment from three dimension with the findings:
1) SiReF is versatile and can be applied to various domains;
2) SiReF is capable of handling the issue of ambiguity in expressions.
3) SiReF can adapt to LLMs with different capabilities.

\textbf{F4} Prompt Engineering plays a critical role for LLMs in different scenario. 
Is there any prompt engineering can improve the effctivness of SiReF on information extraction from HLDs?
We carry out experiment and give three kinds of usefule findings:
1) Numerical precision enhancement. In numerically sensitive scenarios, such as financial reports, SiReF can help in extracting more accurate numerical values.
2) Keyword Completion. With the help from the metadata of document, SiReF can better undertand user-given keyword.
3) Few-Shot Learning. The shot is useful to help SiReF undertand the task. But too many shot may decrease the performance of LLMs to specific tasks. 

To answer some questions, we need to use the globally optimal implementation of SiReF. So we first introduce the SiReF and integrate globally optimal findings for the findings in each module (in \autoref{sc:framework}). And introduce the dataset and metric used for experiments (in \autoref{subsec: dataset}). Next, we show the overall potential of SiReF. Then, we will present experimental analyses of various aspects of SiReF in each section.

\section{Prepared Work}\label{sc:framework}
 \subsection{Automated Information Extraction}  
In order to enable LLMs to handle HLDs, we propose a framework called Automated Information Extraction (AIE), specifically designed for extracting information from HLDs.
The AIE framework consists of four modules: Segmentation, Retrieval, Summarization, and Extraction, as shown in \autoref{fig:framework}.
AIE first segments documents into manageable pieces for LLMs, then retrieves the most relevant segments related to the keyword based on embedding, followed by summarizing the retrieved segments to compress and consolidate critical information and finally extracting the keyword-corresponding value from the generated summary.
This is a feasible framework, there are many implementations for each module.
In the following text, we will introduce each module and provide the optimal implementation based on our experimental analysis.

\subsection{Segmentation}
Despite LLMs vastly improving sequence length handling compared to traditional models like text-davinci-003, which can process 4,097 tokens, HLDs often contain even more tokens.
To address this challenge, we employ this module to split documents into segments that LLMs can handle.
\autoref{fig:framework} demonstrates this module's three steps: Serialization, Split, and Merge.

\textbf{Serialization:} Serialize tables into text.
In hybrid documents, most information is found within tables.
However, LLMs are designed for processing text, so we need a method to convert tables into a textual format.
After comparing various methods (as discussed in \autoref{subsec: Analysis of Table Serialization Formats}), we discover that the \textit{PLAIN} serialization method, which separates cells with spaces and rows with newline characters, provides a simple yet effective way to represent table data in financial reports.

\textbf{Split:} Split long elements. 
In HLDs, there may be exceptionally long elements, such as large tables and extensive paragraphs, which far exceed the processing capacity of LLMs.
To enable LLMs to handle these elements and avoid information loss, we easily divide the overlong paragraphs and tables into small sub-elements based on the LLM's maximum sequence length.

\textbf{Merge:} Merge adjacent elements as segments. 
The primary reason for merging is to maintain semantic relationships between adjacent elements. 
Most elements have a small token count (tens of tokens), which makes merging feasible. 
To achieve this, we concatenate adjacent small elements until the segment length limit is reached

\subsection{Retrieval}
Long documents contain many tokens, leading to a large number of document segments. Processing all segments would significantly increase LLMs invocations and introduce irrelevant information, potentially affecting extraction accuracy. Therefore, we adopt an embedding-based retrieval strategy \cite{li2021embedding} to select the most relevant segments. We calculate the similarity between each document segment and the keyword based on their embeddings and retrieve the top-ranked segments with the highest similarity scores.

To obtain embeddings, we use the Sentence-Transformer model~\cite{reimers2019sentence} in this module. Due to its sequence length limitations smaller than LLMs, document segments are divided into multiple slices.
The similarity between each slice and the keyword is calculated, and the maximum similarity value among the slices within a segment is considered as the similarity between the segment and the keyword.

\subsection{Summarization}
The content related to a keyword is often distributed across various segments. 
To effectively extract and concentrate information, the summarization module leverages LLMs to generate a summary containing relevant information from selected segments.

Since LLMs can only process one segment per invocation, a strategy is needed to connect different segments effectively.
After comparing two common strategies (as discussed in~\autoref{sc:ana strategy}), we apply the \textbf{Refine Strategy} to maintain an evolving summary, updated with information from each segment.

The Refine Strategy process comprises two main steps, depicted in the Summarization module of~\autoref{fig:framework}.
First, the \textit{Question prompt} generates an initial summary from the first segment, guiding LLMs to extract relevant information. Next, the \textit{Refine prompt} updates the summary by incorporating information from the remaining segments. 

\subsection{Extraction}
After the summarization, we obtain a summary that contains the keyword's value along with a considerable amount of irrelevant information.
To eliminate irrelevant information and improve the accuracy and efficiency of downstream tasks, it becomes essential to extract the numerical value.

As shown in the Extraction module of \autoref{fig:framework}, LLMs are utilized to extract the value from the summary. By leveraging the \textit{Extraction Prompt}, LLMs can accurately achieve this goal.

\subsection{Prompt Engineering}
\label{sc:Prompt Engineering}
Prompt Engineering plays an important role in enhancing LLMs' ability~\cite{huang2022towards}.
There are various types of prompt engineering that can be used in AIE. In this work, considering the characteristics of the task of extracting information from HLDs, we have explored the following three types of prompts.

\textbf{Numerical Precision Enhancement:} In scenarios with more numerical data, we find that LLMs have difficulties in maintaining accurate numerical precision. 
For example, the same keyword could correspond to values with different precision levels, all being correct, but the LLMs might not return the most precise result. 
However, in financial analysis, precision is essential for the work. 
To tackle this issue, precision control instructions are incorporated into the prompt, directing the model to produce precise responses.
After comparing many precision-enhancing method (as discussed in~\autoref{sc:ana_precision}), we combine the use of two methods:
Direct and Shot-Precision.
The Direct method directly informs LLMs of the required precision, while the Shot-Precision method demonstrates how to manage precision through input-output examples.

\textbf{Keyword Completion:} Incomplete keywords provided by users can lead to inaccurate IE. 
For example, users might inquire about \textit{Revenue}, but in financial reports, the same keyword might correspond to multiple entities (such as different subsidiaries or time periods). 
To address this issue, we introduce a keyword completion method. 
In our implementation, we utilize the document's metadata.
According to our analysis (as discussed in~\autoref{sc:ana_keyword}), providing more contextual information can greatly improve the accuracy of AIE.

\textbf{Few-Shot Learning:}
In-context learning greatly influences LLMs' capabilities to understand the given task.
According to our analysis (as discussed in~\autoref{sc:ana_shot}),
we find a single well-designed shot is sufficient to guide LLMs to generate accurate answers. On the contrary, excessive shots may potentially reduce the performance of AIE.

\section{Dataset}

\subsection{Datasets on Three Domains} \label{subsec: dataset}

\begin{table}[htbp]  
    \centering  
    \small  
    \begin{tabular}{l|ccc}  
        \hline  
        Dataset & FINE & WIKIR & MPP \\ \hline   
          
        Max \# tokens & 234,900 & 58,512 & 123,105\\   
          
        Min \# tokens & 13,022 & 13,548 & 3,672\\   
          
        Avg. \# tokens & 59,464.3 & 30,922.1 & 17,553.05 \\ \hline  
    \end{tabular}  
    \caption{Basic statistics for FINE, WIKIR, and MPP datasets.}  
    \label{tab: statistics}
    \vspace{-0.5cm}  
\end{table}  

To assess LLMs' capacity to comprehend HLDs and support future research,
we conduct experiments in three representative domains: financial reports, Wikipedia, and scientific papers. 
We construct a dataset for each domain.
The basic statistics can be found in~\autoref{tab: revenue appearance}.
Among these datasets, the financial dataset is used to analyze the various modules of AIE.
The overall performance is tested on all datasets.
For more details about these three datasets, please refer to~\autoref{se:appendix_datasets}.

In the financial reports domain, we introduce a new dataset called the \textbf{Fi}nancial Reports \textbf{N}umerical \textbf{E}xtraction (\textbf{FINE}), comprising manually extracted KPIs from SEC's EDGAR\footnote{\href{https://www.sec.gov/edgar/}{https://www.sec.gov/edgar/}}. 
Using the financial report as content, financial KPIs and related values are utilized as (key, value) pairs.

In the Wikipedia domain, we select the Wikireading-Recycled (WIKIR) dataset~\cite{DBLP:journals/corr/abs-2011-03228}. 
A Wikipedia page serves as the content, while the corresponding key and value are extracted from Wikidata. 

In the scientific papers domain, we select the MPP (Massive Paper Processing) dataset~\cite{polak2023flexible}.
A scientific paper serves as the content, with chemical materials as the keys and their corresponding cooling rates as the values.

\subsection{Evaluation Metrics}

For the FINE, we use the Relative Error Tolerance Accuracy (RETA) metric, for the two other datasets, we use the Accuracy (Acc) metric.

In FINE, all ground truth values are presented in millions, rounded to two decimal places. However, in financial reports, the numerical precision is not uniform, as the values can be expressed in different units, such as millions or billions. This leads to the same keyword being associated with multiple values of varying precision, making it difficult to evaluate the accuracy of LLMs' predictions.

To address this issue, we use the Relative Error Tolerance Accuracy (RETA) metric, which considers predictions as correct if their relative error falls within a specified tolerance threshold (e.g., RETA X\% means predictions with a relative error of no more than X\% are considered correct).
By setting different RETA levels, we can assess the model's performance according to various practical requirements and gain a comprehensive understanding of its capabilities in IE from financial reports.

However, this issue does not exist in the WIKIR and MPP datasets. 
In the WIKIR dataset, the ground truth is a string. 
In the MPP dataset, the ground truth is a floating-point number, and this floating-point number has no alternative precision representation.

\section{Overall Performance on Three Domains}
AIE is a flexible framework. 
To showcase the overall performance of its various modules, we compare AIE with the naive method on all three datasets. 
The AIE used in this section uses the optimal implementation for each module based on our findings (as discussed in the next sections).
The naive method directly uses LLMs adopted to HLDs.
We take the GPT-3.5 (text-davinci-003) as our primary subject.
For the detailed configurations, please refer to \autoref{sc:exp_setting}.

\begin{figure}[ht]  
    \centering  
    \includegraphics[width=\columnwidth]{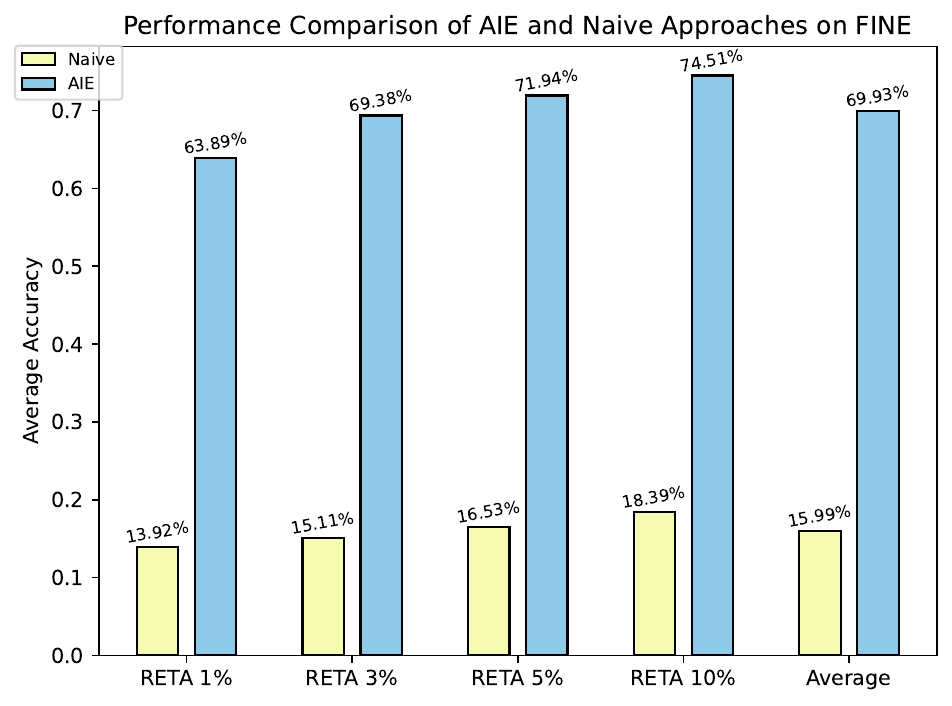}  
    \caption{Comparison of the Naive method and AIE at different RETA levels on FINE.}
    \label{fig: exp_main_result}  
    \vspace{-0.35cm}
\end{figure}

\begin{figure}[ht]  
    \centering  
    \includegraphics[width=\columnwidth]{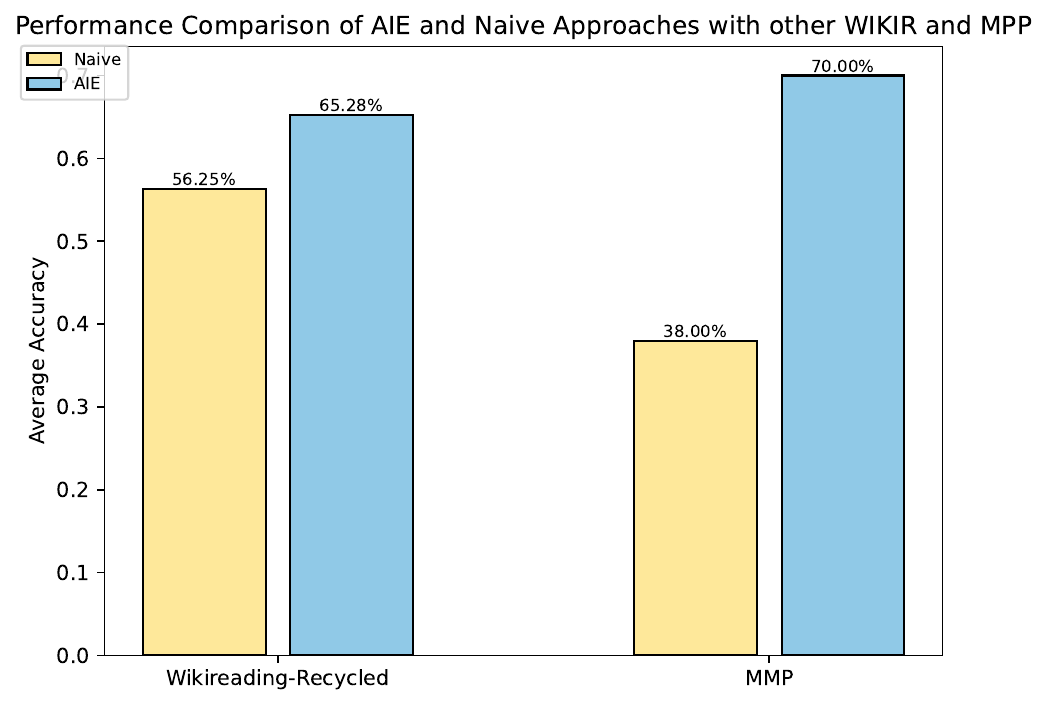}  
    \caption{Comparison of the Naive method and AIE on WIKIR and MPP.}
    \label{fig: new_dataset_result}  
    \vspace{-0.35cm}
\end{figure}

\label{subsec: Comprehensive Results}
The \autoref{fig: exp_main_result} displays the experimental results on FINE. 
It shows the accuracy at different RETA levels, ranging from 1\% to 10\%, and the average accuracy across all RETA settings.
The \autoref{fig: new_dataset_result} displays the experimental results on WIKIR and MPP. 
It shows the average accuracy across all samples.

The experimental results demonstrate that the AIE method outperforms the naive method in all three datasets. The improvement in average accuracy indicates that the AIE method is more effective in extracting relevant information from various HLDs.

In~\autoref{fig: exp_main_result}, as the RETA becomes more stringent, the performance gap between the naive method and AIE becomes larger. This indicates that AIE is capable of delivering more accurate results under stricter evaluation metrics.

\section{Adaptability for LLMs with Different Capabilities}

\begin{figure}[ht]  
    \centering  
    \includegraphics[width=\columnwidth]{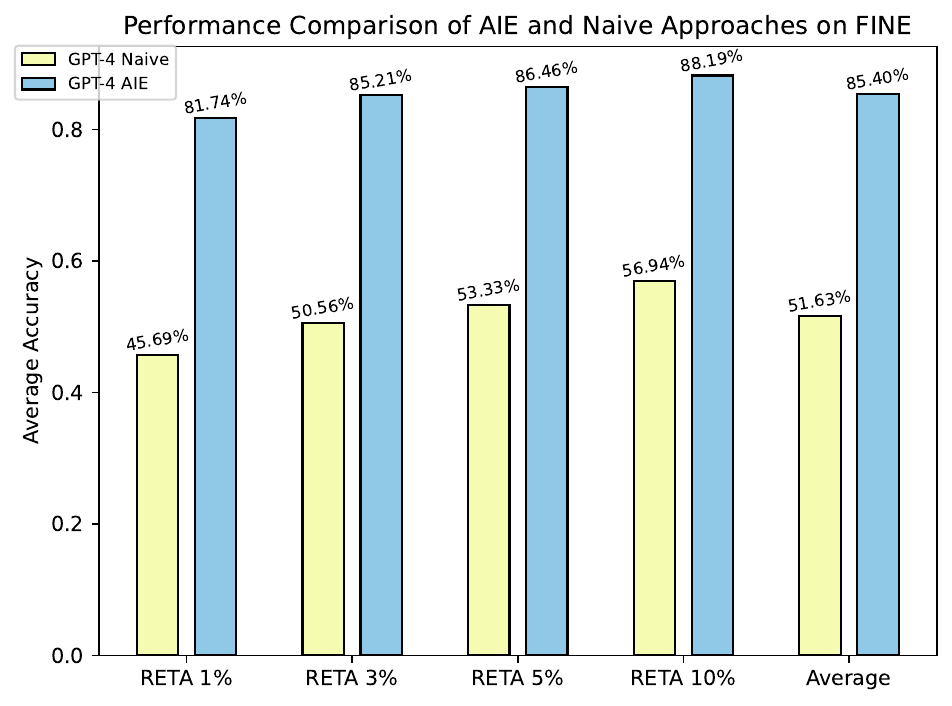}  
    \caption{Comparison of the Naive method and AIE at different RETA levels on GPT-4.}
    \label{fig: result_gpt4}  
    \vspace{-0.35cm}
\end{figure}

To investigate the adaptability of AIE for LLMs with different capabilities, we also conduct experiments on GPT-4. 
For the reason that GPT-4 is currently the most outstanding LLM in terms of comprehensive capabilities.
GPT-4 can handle sequences with a maximum length of 32,768 tokens, while the average length of each sample on WIKIR and MPP datasets does not exceed 32,768 tokens. 
Compared to GPT-4, WIKIR and MPP are not \textbf{long} documents. 
Therefore, we chose the FINE as the experimental subject.

The \autoref{fig: result_gpt4} displays the experimental results of AIE on GPT-4.
From the results, we can see that when using GPT-4, AIE's performance is still better than the naive strategy under different RETA levels.
This demonstrates AIE's adaptability ability on LLMs with different capabilities.

\section{Capability to Handle Ambiguity}
In HDLs, the same concept may have multiple representations, which requires LLMs to have the ability to handle ambiguity. 
To evaluate whether AIE can enhance such ability, we conduct a comparison on two sets of keywords: (\textit{Revenue} vs. \textit{Total Net Sales}) and (\textit{Total Equity} vs. \textit{Total Stockholders' Equity}).
We compare the Relative Percentage Difference (RPD) in average accuracy between the naive method and AIE across various RETA levels. The RPD at a certain RETA level is calculated using the following formula:
$$RPD_{X-Y} = \frac{abs(Acc_X - Acc_Y)}{average(Acc_X, Acc_Y)}$$
where \(Acc_X\) and \(Acc_Y\) represent the average accuracy of two different keywords.

\begin{figure*}[ht]
\centering
\includegraphics[width=0.8\textwidth]{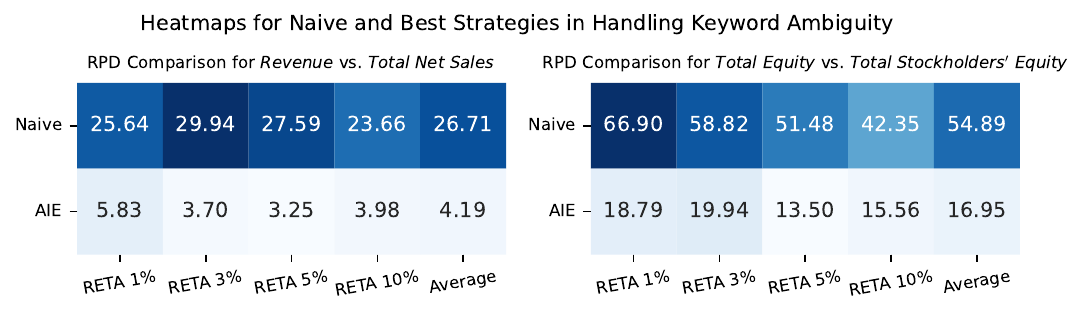}
\caption{Exploring the Capability to Handle Keyword Ambiguity: Comparison of Naive and AIE on RPD}
\label{fig:heatmap_keyword_ambiguity}
\vspace{-0.50cm}
\end{figure*}

The experimental results are presented in~\autoref{fig:heatmap_keyword_ambiguity}.
From the results, we observe that AIE outperforms the naive method across all RETA levels when handling keyword ambiguity.

Specifically, comparing \textit{Revenue} vs. \textit{Total Net Sales}, AIE shows a 22.52\% lower avg. RPD than the naive method. Similarly, for \textit{Total Equity} vs. \textit{Total Stockholders' Equity}, AIE yields a 37.94\% lower avg. RPD than the naive method. For more detailed results, please refer to the~\autoref{sec: Detailed Results of Keyword Ambiguity Experiment}.

\section{Analysis of Table Serialization Formats}
\label{subsec: Analysis of Table Serialization Formats} 
\begin{table}[ht]        
\centering        
\fontsize{8}{10}\selectfont 
\setlength{\tabcolsep}{2pt}
\begin{tabular}{l|cccc|c}        
\hline        
& \textbf{RETA 1\%} & \textbf{RETA 3\%} & \textbf{RETA 5\%} & \textbf{RETA 10\%} & \textbf{Average} \\ \hline    
\textbf{PLAIN} & \textbf{0.6389} & \textbf{0.6938} & \textbf{0.7194} & \textbf{0.7451} & \textbf{0.6993} \\        
\textbf{CSV} & 0.6264 & 0.6889 & 0.7132 & 0.7361 & 0.6911 \\        
\textbf{XML} & 0.3951 & 0.4507 & 0.4729 & 0.5069 & 0.4564 \\       
\textbf{HTM}L & 0.4542 & 0.5000 & 0.5208 & 0.5590 & 0.5085 \\ \hline        
\end{tabular} 
\caption{Accuracy comparison among PLAIN, CSV, XML, and HTML table serialization formats.}
\label{tab:serialization_formats}
\vspace{-0.15cm}
\end{table}

In order to enable LLMs to handle tabular data, we need to use a specific serialization method to represent tables as text. 
There are four common serialization methods: PLAIN, CSV, XML, and HTML.

\textbf{PLAIN} serialization extracts text from table cells, separating adjacent cell content with spaces and using newline characters to separate rows. \textbf{CSV} serialization separates adjacent cells with comma delimiters. \textbf{XML} and \textbf{HTML} serialization formats utilize tags\footnote{XML employs tags such as <table>, <row>, and <cell>, while HTML utilizes tags like <tr> (for table rows) and <td> (for table cells).} to preserve the hierarchical relationships between table elements.

Despite XML and HTML formats retaining hierarchical information, the incorporation of tags results in a higher token count, potentially exceeding the LLMs' maximum sequence length and requiring more frequent table splitting. 
As shown in \autoref{tab:serialization_formats}, the PLAIN and the CSV formats outperform the XML and HTML formats in terms of accuracy, likely due to their concise table representation, which reduces table fragmentation and captures the complete semantic information of the tables.

\section{Analysis of Retrieved Segment Number}
\label{subsec: Impact of Retrieved Segment Quantity on Accuracy}
\begin{table}[ht]      
\centering  
\fontsize{8}{10}\selectfont 
\setlength{\tabcolsep}{2pt} 
\begin{tabular}{l|cccc|c}      
\hline      
 & \textbf{RETA  1\%} & \textbf{RETA  3\%} & \textbf{RETA  5\%} & \textbf{RETA  10\%} & \textbf{Average} \\ \hline      
\textbf{R@1} & 0.4757 & 0.5278 & 0.5444 & 0.5694 & 0.5293 \\ 
\textbf{R@2} & 0.6188 & 0.6736 & 0.6931 & 0.7118 & 0.6743 \\ 
\textbf{R@3} & \textbf{0.6389} & \textbf{0.6938} & \textbf{0.7194} & \textbf{0.7451} & \textbf{0.6993} \\      
\textbf{R@5} & 0.6160 & 0.6799 & 0.7062 & 0.7306 & 0.6832 \\     
\textbf{R@7} & 0.5917 & 0.6521 & 0.6722 & 0.7090 & 0.6563 \\       
\textbf{No R} & 0.3757 & 0.4986 & 0.5201 & 0.5514 & 0.4865 \\ \hline      
\end{tabular}      
\caption{Accuracy comparison for different retrieval quantities (R@n) across various RETA levels.} 
\label{tab:retrieval_quantity}
\vspace{-0.3cm}
\end{table}

In this section, we investigate the effect of the number of retrieved segments on the performance of our framework.
\autoref{tab:retrieval_quantity} shows the accuracy for different retrieval quantities, where R@n represents the number of top-ranked segments retrieved.

The results reveal that the highest accuracy across all RETA levels is achieved when the retrieval quantity is set to 3 (R@3). 
Analyzing the trend, we can observe that the accuracy increases as the retrieval quantity goes from 1 to 3, demonstrating the benefits of retrieving more segments to capture additional information. 
However, as the retrieval quantity increases beyond 3, the accuracy declines. 
This suggests that including too many segments may introduce noise or irrelevant information, which adversely affects performance. 

\section{Analysis of Summarization Strategies}
\label{sc:ana strategy}

\begin{figure}[ht]  
    \centering  
    \includegraphics[width=0.8\linewidth]{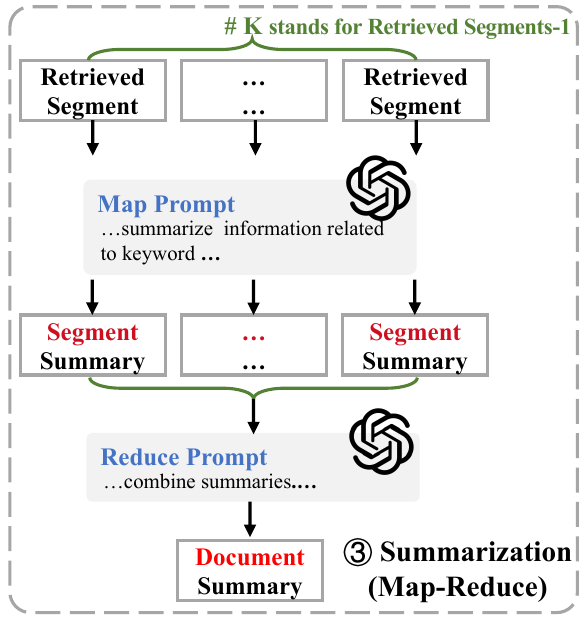}  
    \caption{Illustration of the Map-Reduce Strategy, comprising two stages: Map, generating individual segment summaries, and Reduce, combining these summaries to form a single document summary.}  
    \label{fig:map_reduce_strategy}  
    \vspace{-0.5cm}
\end{figure}

In order to extract information from multiple retrieved segments, several popular strategies are available. 
Besides the Refine Strategy, another widely used strategy is the Map-Reduce Strategy, which is known for its parallel processing capabilities. 
As illustrated in~\autoref{fig:map_reduce_strategy}, the Map-Reduce Strategy aims to combine summaries from document segments, comprising two stages: Map and Reduce. 
In the Map stage, LLMs generate a segment summary for each document segment in parallel. 
During the Reduce stage, LLMs consolidate all the segment summaries to form a cohesive document summary. 

\begin{table*}[htbp]      
\centering     
\small
\begin{tabular}{l|cccc|cc}      
\hline      
& \textbf{RETA 1\%} & \textbf{RETA 3\%} & \textbf{RETA 5\%} & \textbf{RETA 10\%} & \textbf{Average} & \textbf{Time (s\textbackslash sample)} \\ \hline      
\textbf{Map-Reduce} & 0.5375 & 0.5729 & 0.5958 & 0.6299 & 0.5840 &  \textbf{13.34} \\ 
\textbf{Refine} & \textbf{0.6389} & \textbf{0.6938} & \textbf{0.7194} &\textbf{0.7451} & \textbf{0.6993} & 16.36 \\ \hline
\end{tabular}    
\caption{Accuracy comparison between Map-Reduce and Refine strategies across various RETA levels.}   
\label{tab:summarization_strategies}
\vspace{-0.5cm}
\end{table*}  

As shown in~\autoref{tab:summarization_strategies}, the Refine Strategy consistently outperforms the Map-Reduce Strategy in terms of accuracy across all RETA levels. 
However, it is essential to consider the trade-off between accuracy and efficiency when selecting a summarization strategy for a given application. 
The Map-Reduce Strategy offers the advantage of parallel processing, making it a better choice for situations where processing speed is of higher importance.

\section{Analysis of Numerical Precision Enhancement}
\label{sc:ana_precision}
In order to enable LLMs to extract more accurate numerical values, we design various numerical precision enhancement prompts.
To assess the performance of these prompts, we conducted a comparative experiment under finer RETA levels.

\textbf{TD-O}: Task description only.
\textbf{TD-R}: Naive prompt with precision requirements.
\textbf{TD-S}: Naive prompt with input-output example.
\textbf{TD-RS}: Naive prompt, precision requirements, and input-output example.
\textbf{TD-SP}: Naive prompt with precision-inclusive input-output example.
\textbf{TD-RSP}: Naive prompt, precision requirements, and precision-inclusive input-output example.
See~\autoref{app: Numerical Precision Enhancement Prompts} for details of these prompts.

\begin{table}[ht]  
  \centering  
  \fontsize{9}{11}\selectfont  
  \setlength{\tabcolsep}{3pt}  
  \begin{tabular}{l|cccc|c}  
    \hline  
    & \multicolumn{4}{c|}{\textbf{RETA}} & \\  
    & \textbf{0\%} & \textbf{0.001\%} & \textbf{0.01\%} & \textbf{0.1\%} & \textbf{Average} \\  
    \hline  
    \textbf{TD-O} & 0.4917 & 0.4937 & 0.5187 & 0.5750 & 0.5198 \\  
    \textbf{TD-R} & 0.3479 & 0.3479 & 0.3597 & 0.4083 & 0.3660 \\  
    \textbf{TD-S} & 0.4111 & 0.4153 & 0.4493 & 0.5438 & 0.4549 \\  
    \textbf{TD-RS} & 0.4403 & 0.4438 & 0.4722 & 0.5396 & 0.4740 \\  
    \textbf{TD-SP} & 0.5278 & 0.5299 & 0.5479 & 0.5882 & 0.5484 \\  
    \textbf{TD-RSP} & \textbf{0.5646} & \textbf{0.5660} & \textbf{0.5750} & \textbf{0.5938} & \textbf{0.5748} \\  
    \hline  
  \end{tabular}  
  \caption{Accuracy comparison for different prompts aimed at enhancing numerical precision.}  
  \label{tab:precision_comparison}  

\end{table}  

From~\autoref{tab:precision_comparison}, we observe the following:
\textbf{1)} The TD-RSP strategy achieves the highest accuracy across all fine-grained RETA levels, indicating its effectiveness in enhancing the numerical precision of extracted values.  
\textbf{2)} The performance of TD-R, TD-S, and TD-RS strategies is inferior to that of TD-O. This may suggest that improperly designed or insufficient precision prompts could act as a distractor, hindering its ability to focus on improving numerical accuracy.  

\section{Analysis of Keyword Completion}
\label{sc:ana_keyword}
To analyze the effectiveness of keyword completion in improving LLMs' performance, we conducted an experiment with various settings.

\textbf{K}: Only provide keyword names, such as ``Net Income'', ``Revenue'', etc.  
\textbf{K\_C}: Provide keyword names and company names, such as ``Net Income of Nike''.  
\textbf{K\_T}: Provide keyword names and time, such as ``Net Income of 2022Q4''.  
\textbf{K\_T\_C}: Provide keyword names, time, and company names, such as ``Net Income of Nvidia 2022Q4''.

\begin{table}[ht]      
\centering
\setlength{\tabcolsep}{3pt} 
\fontsize{8}{10}\selectfont 
\setlength{\tabcolsep}{1pt}
\begin{tabular}{l|cccc|c}      
\hline      
& \textbf{RETA 1\%} & \textbf{RETA 3\%} & \textbf{RETA 5\%} & \textbf{RETA 10\%} & \textbf{Average} \\ \hline      
\textbf{K} & 0.3403 & 0.3917 & 0.4076 & 0.4292 & 0.3922 \\ 
\textbf{K\_C} & 0.4681 & 0.5167 & 0.5361 & 0.5604 & 0.5203 \\ 
\textbf{K\_T} & 0.4785 & 0.5396 & 0.5500 & 0.5736 & 0.5354 \\ 
\textbf{K\_T\_C} & \textbf{0.6389} & \textbf{0.6938} & \textbf{0.7194} & \textbf{0.7451} & \textbf{0.6993} \\ \hline      
\end{tabular}      
\caption{Accuracy comparison for different keyword completion settings across various RETA levels.}

\label{tab:keyword_completion_comparison}
\vspace{-0.35cm}
\end{table}  

As shown in~\autoref{tab:keyword_completion_comparison}, we find that the performance of \textit{K\_C}, \textit{K\_T}, and \textit{K\_T\_C} strategies is better than that of \textit{K}, with \textit{K\_T\_C} achieving the best results. This indicates that keyword completion is useful in improving LLMs' accuracy. By providing more specific information, such as company names and time periods, the model can better understand the context and generate more accurate responses, leading to an overall improvement in performance.

\section{Analysis of Shot Number}
\label{sc:ana_shot}
Few-shot learning is an important ability of LLMs. To investigate the impact of the number of shots on AIE's performance, we conducted an experiment with different numbers of shots, ranging from 0 to 3.

\begin{table}[ht]  
\centering  
\fontsize{8}{10}\selectfont  
\setlength{\tabcolsep}{2pt} 
\begin{tabular}{l|cccc|c}
\hline  
& \textbf{RETA 1\%} & \textbf{RETA 3\%} & \textbf{RETA 5\%} &\textbf{RETA 10\%} & \textbf{Average} \\ \hline  
\textbf{0-shot} & 0.4799 & 0.5229 & 0.5354 & 0.5472 & 0.5214 \\ 
\textbf{1-shot} & \textbf{0.6389} & \textbf{0.6938} & \textbf{0.7194} & \textbf{0.7451} & \textbf{0.6993} \\ 
\textbf{2-shot} & 0.6227 & 0.6803 & 0.6966 & 0.7231 & 0.6807 \\  
\textbf{3-shot} & 0.6181 & 0.6806 & 0.7007 & 0.7174 & 0.6792 \\ \hline  
\end{tabular}  
\caption{Accuracy comparison for different numbers of shots across various RETA levels.}  
\label{tab:number_of_shots_comparison}  
\end{table}  

As shown in \autoref{tab:number_of_shots_comparison}, the 1-shot setting achieves the highest accuracy across all RETA levels. The performance of 2-shot and 3-shot settings is slightly lower than that of the 1-shot setting but still better than the 0-shot setting. This indicates that a single well-designed example can effectively guide LLMs to generate more accurate responses. However, the slight decrease in performance with additional examples could be attributed to the increased complexity of the input or potential inconsistencies among multiple examples, which may confuse the model rather than provide more guidance.

Based on this experiment, we recommend carefully determining the number of shots when using LLMs for information extraction.
Although providing more shots may still be helpful, it is essential to ensure their consistency and relevance to avoid potential confusion and maintain optimal performance.

\section{Discussion}
In addition to our extensive exploration experiments with AIE, we also eliminate the interference of pre-trained datasets on the experiments \autoref{sc:pre-train}, ensuring the reliability of our results. We find that using AIE significantly reduces the number of LLM invocations \autoref{sc:ana_cost}. Furthermore, it is essential to use both tabular and textual data simultaneously in HLDs \autoref{sc:nec_both_data}.

\section{Related Work}
\textbf{Information extraction.}
Early IE methods predominantly relied on rule-based approaches.
\cite{sheikh2012rule, farmakiotou2000rule} proposed rule-based Named Entity Recognition (NER)  models that utilize domain-specific characteristics.
Among them, some approaches concentrate on tables \cite{brito2019hybrid}, losing crucial textual information.
Recently, models based on Machine Learning emerged, in which bidirectional RNN classifier \cite{ma2020spot} is employed for learning tables, and BERT \cite{hillebrand2022kpi} for text respectively. 
FinQA \cite{chen2022finqa}, TAT-QA \cite{zhu2021tatqa}, and MULTIHIERTT \cite{zhao-etal-2022-multihiertt} learn HLDs for Question Answer (QA) task, which focus on parts of the reports rather than the whole context.

\textbf{LLMs.}
In our research, we primarily focus on leveraging the capabilities of LLMs across three distinct tasks.
1) Long document processing, helping LLMs exceed their maximum input length limit \cite{liang2023unleashing}.
2) IE, particularly value extraction, where LLMs have shown proficiency in the domains such as IE \cite{li2023evaluating,wei2023zero}, which includes NER \cite{gupta2021context,wang2023gpt}, Relation Extraction (RE) \cite{wan2023gpt,xu2023unleash}, and Knowledge Graph Extraction \cite{shi2023chatgraph}.
\cite{polak2023flexible,arora2023language} have successfully demonstrated the extraction of key-value pairs from the text content of academic papers and HTML respectively, thereby substantiating the dependability of LLMs for value extraction. 
3) Tabular reasoning, where LLMs have demonstrated considerable ability to perform intricate reasoning tasks with structured data \cite{chen2023large, ye2023large}.

\section{Conclusion}

In order to enable LLMs to extract information from HLDs, we propose an \textbf{A}utomated \textbf{I}nformation \textbf{E}xtraction (\textbf{AIE}) framework, comprising four modules: Segmentation, Retrieval, Summarization, and Extraction.
To analyze the various modules of AIE, we construct a dataset from publicly available financial reports, called \textbf{Fi}nancial Reports \textbf{N}umerical \textbf{E}xtraction (\textbf{FINE}).
Based on the FINE, experiments offer a detailed explanation of the impact of each module on AIE's ability to extract information from HLDs.
We also validate the overall performance of AIE on three different domains: scientific papers, Wikipedia, and financial reports.
Experimental results show that AIE significantly improves the ability of LLMs to handle HLDs.

\section*{Limitations}

Despite the substantial enhancement achieved by LLMs through the utilization of AIE, certain limitations persist.
\begin{enumerate}
    \item Model ability limitation: This work effectively demonstrates LLMs' ability to extract information from HLDs. However, further evaluation of their capabilities in other aspects, such as formula inferencing, generating abstracts, and keyword extraction, remains necessary.

    \item Multimodal limitations: AIE can effectively extract information from documents containing a mix of textual and tabular data. However, its capabilities in handling other types of content within documents, such as images, diagrams, or complex visualizations, have not been evaluated. In many real-world scenarios, HLDs may contain rich multi-modal information that could be crucial for making informed decisions.
    
    \item Cost constraints: The GPT-3.5 and GPT-4 used in the experiments incur computational costs. For some practical applications, AIE may not be the most cost-effective method. 
\end{enumerate}

\bibliography{anthology}

\begin{thebibliography}{32}
\expandafter\ifx\csname natexlab\endcsname\relax\def\natexlab#1{#1}\fi

\bibitem[{Arora et~al.(2023)Arora, Yang, Eyuboglu, Narayan, Hojel, Trummer, and R{\'e}}]{arora2023language}
Simran Arora, Brandon Yang, Sabri Eyuboglu, Avanika Narayan, Andrew Hojel, Immanuel Trummer, and Christopher R{\'e}. 2023.
\newblock Language models enable simple systems for generating structured views of heterogeneous data lakes.
\newblock \emph{arXiv preprint arXiv:2304.09433}.

\bibitem[{Brito et~al.(2019)Brito, Sifa, Bauckhage, Loitz, Lohmeier, and P{\"u}nt}]{brito2019hybrid}
Eduardo Brito, Rafet Sifa, Christian Bauckhage, R{\"u}diger Loitz, Uwe Lohmeier, and Christin P{\"u}nt. 2019.
\newblock A hybrid ai tool to extract key performance indicators from financial reports for benchmarking.
\newblock In \emph{Proceedings of the ACM Symposium on Document Engineering 2019}, pages 1--4.

\bibitem[{Chen(2023)}]{chen2023large}
Wenhu Chen. 2023.
\newblock \href {http://arxiv.org/abs/2210.06710} {Large language models are few(1)-shot table reasoners}.

\bibitem[{Chen et~al.(2022)Chen, Chen, Smiley, Shah, Borova, Langdon, Moussa, Beane, Huang, Routledge, and Wang}]{chen2022finqa}
Zhiyu Chen, Wenhu Chen, Charese Smiley, Sameena Shah, Iana Borova, Dylan Langdon, Reema Moussa, Matt Beane, Ting-Hao Huang, Bryan Routledge, and William~Yang Wang. 2022.
\newblock \href {http://arxiv.org/abs/2109.00122} {Finqa: A dataset of numerical reasoning over financial data}.

\bibitem[{da~Silva et~al.(2023)da~Silva, Milios, and de~Oliveira}]{DBLP:journals/iwc/SilvaMO23}
Sherlon~Almeida da~Silva, Evangelos~E. Milios, and Maria Cristina~Ferreira de~Oliveira. 2023.
\newblock \href {https://doi.org/10.1093/iwc/iwad019} {Evaluating visual analytics for relevant information retrieval in document collections}.
\newblock \emph{Interact. Comput.}, 35(2):247--261.

\bibitem[{Dwojak et~al.(2020)Dwojak, Pietruszka, Borchmann, Chledowski, and Gralinski}]{DBLP:journals/corr/abs-2011-03228}
Tomasz Dwojak, Michal Pietruszka, Lukasz Borchmann, Jakub Chledowski, and Filip Gralinski. 2020.
\newblock From dataset recycling to multi-property extraction and beyond.
\newblock \emph{CoRR}, abs/2011.03228.

\bibitem[{Farmakiotou et~al.(2000)Farmakiotou, Karkaletsis, Koutsias, Sigletos, Spyropoulos, and Stamatopoulos}]{farmakiotou2000rule}
Dimitra Farmakiotou, Vangelis Karkaletsis, John Koutsias, George Sigletos, Constantine~D Spyropoulos, and Panagiotis Stamatopoulos. 2000.
\newblock Rule-based named entity recognition for greek financial texts.
\newblock In \emph{Proceedings of the Workshop on Computational lexicography and Multimedia Dictionaries (COMLEX 2000)}, pages 75--78.

\bibitem[{Gil(2023)}]{DBLP:journals/csr/Gil23}
Jorge~Mart{\'{\i}}nez Gil. 2023.
\newblock \href {https://doi.org/10.1016/j.cosrev.2023.100552} {A survey on legal question-answering systems}.
\newblock \emph{Comput. Sci. Rev.}, 48:100552.

\bibitem[{Gupta et~al.(2021)Gupta, Verma, Kumar, Mishra, Agrawal, Badugu, and Bhatt}]{gupta2021context}
Himanshu Gupta, Shreyas Verma, Tarun Kumar, Swaroop Mishra, Tamanna Agrawal, Amogh Badugu, and Himanshu~Sharad Bhatt. 2021.
\newblock Context-ner: Contextual phrase generation at scale.
\newblock \emph{arXiv preprint arXiv:2109.08079}.

\bibitem[{Hewlett et~al.(2016)Hewlett, Lacoste, Jones, Polosukhin, Fandrianto, Han, Kelcey, and Berthelot}]{DBLP:journals/corr/HewlettLJPFHKB16}
Daniel Hewlett, Alexandre Lacoste, Llion Jones, Illia Polosukhin, Andrew Fandrianto, Jay Han, Matthew Kelcey, and David Berthelot. 2016.
\newblock Wikireading: {A} novel large-scale language understanding task over wikipedia.
\newblock \emph{CoRR}, abs/1608.03542.

\bibitem[{Hillebrand et~al.(2022)Hillebrand, Deu{\ss}er, Dilmaghani, Kliem, Loitz, Bauckhage, and Sifa}]{hillebrand2022kpi}
Lars Hillebrand, Tobias Deu{\ss}er, Tim Dilmaghani, Bernd Kliem, R{\"u}diger Loitz, Christian Bauckhage, and Rafet Sifa. 2022.
\newblock Kpi-bert: A joint named entity recognition and relation extraction model for financial reports.
\newblock In \emph{2022 26th International Conference on Pattern Recognition (ICPR)}, pages 606--612. IEEE.

\bibitem[{Huang and Chang(2022)}]{huang2022towards}
Jie Huang and Kevin Chen-Chuan Chang. 2022.
\newblock Towards reasoning in large language models: A survey.
\newblock \emph{arXiv preprint arXiv:2212.10403}.

\bibitem[{Kojima et~al.(2023)Kojima, Gu, Reid, Matsuo, and Iwasawa}]{kojima2023large}
Takeshi Kojima, Shixiang~Shane Gu, Machel Reid, Yutaka Matsuo, and Yusuke Iwasawa. 2023.
\newblock \href {http://arxiv.org/abs/2205.11916} {Large language models are zero-shot reasoners}.

\bibitem[{Li et~al.(2023)Li, Fang, Yang, Wang, Ye, Zhao, and Zhang}]{li2023evaluating}
Bo~Li, Gexiang Fang, Yang Yang, Quansen Wang, Wei Ye, Wen Zhao, and Shikun Zhang. 2023.
\newblock Evaluating chatgpt's information extraction capabilities: An assessment of performance, explainability, calibration, and faithfulness.
\newblock \emph{arXiv preprint arXiv:2304.11633}.

\bibitem[{Li et~al.(2021)Li, Lv, Jin, Lin, Yang, Zeng, Wu, and Ma}]{li2021embedding}
Sen Li, Fuyu Lv, Taiwei Jin, Guli Lin, Keping Yang, Xiaoyi Zeng, Xiao-Ming Wu, and Qianli Ma. 2021.
\newblock Embedding-based product retrieval in taobao search.
\newblock In \emph{Proceedings of the 27th ACM SIGKDD Conference on Knowledge Discovery \& Data Mining}, pages 3181--3189.

\bibitem[{Liang et~al.(2023)Liang, Wang, Huang, Wu, Wu, Lu, Ma, and Li}]{liang2023unleashing}
Xinnian Liang, Bing Wang, Hui Huang, Shuangzhi Wu, Peihao Wu, Lu~Lu, Zejun Ma, and Zhoujun Li. 2023.
\newblock Unleashing infinite-length input capacity for large-scale language models with self-controlled memory system.
\newblock \emph{arXiv preprint arXiv:2304.13343}.

\bibitem[{Ma et~al.(2020)Ma, Pomerville, Di, and Nourbakhsh}]{ma2020spot}
Zhiqiang Ma, Steven Pomerville, Mingyang Di, and Armineh Nourbakhsh. 2020.
\newblock Spot: A tool for identifying operating segments in financial tables.
\newblock In \emph{Proceedings of the 43rd International ACM SIGIR Conference on Research and Development in Information Retrieval}, pages 2157--2160.

\bibitem[{Mustafa et~al.(2023)Mustafa, Rauf, Al{-}Shamayleh, Sulaiman, Alrawagfeh, Afzal, and Akhunzada}]{DBLP:journals/access/MustafaRASAAA23}
Ghulam Mustafa, Abid Rauf, Ahmad~Sami Al{-}Shamayleh, Muhammad Sulaiman, Wagdi Alrawagfeh, Muhammad~Tanvir Afzal, and Adnan Akhunzada. 2023.
\newblock \href {https://doi.org/10.1109/ACCESS.2023.3292248} {Optimizing document classification: Unleashing the power of genetic algorithms}.
\newblock \emph{{IEEE} Access}, 11:83136--83149.

\bibitem[{Polak et~al.(2023)Polak, Modi, Latosinska, Zhang, Wang, Wang, Hazra, and Morgan}]{polak2023flexible}
Maciej~P Polak, Shrey Modi, Anna Latosinska, Jinming Zhang, Ching-Wen Wang, Shanonan Wang, Ayan~Deep Hazra, and Dane Morgan. 2023.
\newblock Flexible, model-agnostic method for materials data extraction from text using general purpose language models.
\newblock \emph{arXiv preprint arXiv:2302.04914}.

\bibitem[{Reimers and Gurevych(2019)}]{reimers2019sentence}
Nils Reimers and Iryna Gurevych. 2019.
\newblock Sentence-bert: Sentence embeddings using siamese bert-networks.
\newblock \emph{arXiv preprint arXiv:1908.10084}.

\bibitem[{Sheikh and Conlon(2012)}]{sheikh2012rule}
Mahmudul Sheikh and Sumali Conlon. 2012.
\newblock A rule-based system to extract financial information.
\newblock \emph{Journal of Computer Information Systems}, 52(4):10--19.

\bibitem[{Shi et~al.(2023)Shi, Ma, Zhong, Mai, Li, Liu, and Huang}]{shi2023chatgraph}
Yucheng Shi, Hehuan Ma, Wenliang Zhong, Gengchen Mai, Xiang Li, Tianming Liu, and Junzhou Huang. 2023.
\newblock Chatgraph: Interpretable text classification by converting chatgpt knowledge to graphs.
\newblock \emph{arXiv preprint arXiv:2305.03513}.

\bibitem[{Wan et~al.(2023)Wan, Cheng, Mao, Liu, Song, Li, and Kurohashi}]{wan2023gpt}
Zhen Wan, Fei Cheng, Zhuoyuan Mao, Qianying Liu, Haiyue Song, Jiwei Li, and Sadao Kurohashi. 2023.
\newblock Gpt-re: In-context learning for relation extraction using large language models.
\newblock \emph{arXiv preprint arXiv:2305.02105}.

\bibitem[{Wang et~al.(2023{\natexlab{a}})Wang, Sun, Li, Ouyang, Wu, Zhang, Li, and Wang}]{wang2023gpt}
Shuhe Wang, Xiaofei Sun, Xiaoya Li, Rongbin Ouyang, Fei Wu, Tianwei Zhang, Jiwei Li, and Guoyin Wang. 2023{\natexlab{a}}.
\newblock Gpt-ner: Named entity recognition via large language models.
\newblock \emph{arXiv preprint arXiv:2304.10428}.

\bibitem[{Wang et~al.(2023{\natexlab{b}})Wang, Wei, Schuurmans, Le, Chi, Narang, Chowdhery, and Zhou}]{wang2023selfconsistency}
Xuezhi Wang, Jason Wei, Dale Schuurmans, Quoc Le, Ed~Chi, Sharan Narang, Aakanksha Chowdhery, and Denny Zhou. 2023{\natexlab{b}}.
\newblock \href {http://arxiv.org/abs/2203.11171} {Self-consistency improves chain of thought reasoning in language models}.

\bibitem[{Wei et~al.(2023{\natexlab{a}})Wei, Wang, Schuurmans, Bosma, Ichter, Xia, Chi, Le, and Zhou}]{wei2023chainofthought}
Jason Wei, Xuezhi Wang, Dale Schuurmans, Maarten Bosma, Brian Ichter, Fei Xia, Ed~Chi, Quoc Le, and Denny Zhou. 2023{\natexlab{a}}.
\newblock \href {http://arxiv.org/abs/2201.11903} {Chain-of-thought prompting elicits reasoning in large language models}.

\bibitem[{Wei et~al.(2023{\natexlab{b}})Wei, Cui, Cheng, Wang, Zhang, Huang, Xie, Xu, Chen, Zhang et~al.}]{wei2023zero}
Xiang Wei, Xingyu Cui, Ning Cheng, Xiaobin Wang, Xin Zhang, Shen Huang, Pengjun Xie, Jinan Xu, Yufeng Chen, Meishan Zhang, et~al. 2023{\natexlab{b}}.
\newblock Zero-shot information extraction via chatting with chatgpt.
\newblock \emph{arXiv preprint arXiv:2302.10205}.

\bibitem[{Xu et~al.(2023)Xu, Zhu, Wang, and Zhang}]{xu2023unleash}
Xin Xu, Yuqi Zhu, Xiaohan Wang, and Ningyu Zhang. 2023.
\newblock How to unleash the power of large language models for few-shot relation extraction?
\newblock \emph{arXiv preprint arXiv:2305.01555}.

\bibitem[{Ye et~al.(2023)Ye, Hui, Yang, Li, Huang, and Li}]{ye2023large}
Yunhu Ye, Binyuan Hui, Min Yang, Binhua Li, Fei Huang, and Yongbin Li. 2023.
\newblock Large language models are versatile decomposers: Decompose evidence and questions for table-based reasoning.
\newblock \emph{arXiv preprint arXiv:2301.13808}.

\bibitem[{Zhao et~al.(2022)Zhao, Li, Li, and Zhang}]{zhao-etal-2022-multihiertt}
Yilun Zhao, Yunxiang Li, Chenying Li, and Rui Zhang. 2022.
\newblock \href {https://doi.org/10.18653/v1/2022.acl-long.454} {{M}ulti{H}iertt: Numerical reasoning over multi hierarchical tabular and textual data}.
\newblock In \emph{Proceedings of the 60th Annual Meeting of the Association for Computational Linguistics (Volume 1: Long Papers)}, pages 6588--6600, Dublin, Ireland. Association for Computational Linguistics.

\bibitem[{Zhou et~al.(2022)Zhou, Sch{\"a}rli, Hou, Wei, Scales, Wang, Schuurmans, Bousquet, Le, and Chi}]{zhou2022least}
Denny Zhou, Nathanael Sch{\"a}rli, Le~Hou, Jason Wei, Nathan Scales, Xuezhi Wang, Dale Schuurmans, Olivier Bousquet, Quoc Le, and Ed~Chi. 2022.
\newblock Least-to-most prompting enables complex reasoning in large language models.
\newblock \emph{arXiv preprint arXiv:2205.10625}.

\bibitem[{Zhu et~al.(2021)Zhu, Lei, Huang, Wang, Zhang, Lv, Feng, and Chua}]{zhu2021tatqa}
Fengbin Zhu, Wenqiang Lei, Youcheng Huang, Chao Wang, Shuo Zhang, Jiancheng Lv, Fuli Feng, and Tat-Seng Chua. 2021.
\newblock \href {http://arxiv.org/abs/2105.07624} {Tat-qa: A question answering benchmark on a hybrid of tabular and textual content in finance}.

\end{thebibliography}

\appendix
\section{Details of Datasets}
\label{se:appendix_datasets}
\subsection{FINE}
To the best of our knowledge, there is no suitable HLD dataset in the domain of financial reports. 
So we introduce the \textbf{Fi}nancial Reports \textbf{N}umerical \textbf{E}xtraction (\textbf{FINE}) dataset, comprising manually extracted KPIs from SEC's EDGAR\footnote{\href{https://www.sec.gov/edgar/}{https://www.sec.gov/edgar/}}.
We collect reports from 18 companies across four sectors for a 4-year fiscal period (2019-2022). 
Within a fiscal year, a company's financial reports consist of three quarterly financial reports and one annual financial report.
These companies are categorized into four groups based on their operational domains: technology, retail, financial services, and food and beverage.
We identify 9 commonly used crucial KPIs that exemplify the ambiguous, HLDs characteristics of financial reports.
In FINE, ground truth is represented as tuples of four elements: (company, time, keyword, value)\footnote{
One tuple denotes the value corresponding to a specific keyword for a given company at a specified time. 
For example, (COMPANY, three months ended 2022.12.31, Revenue, 12345.00) indicates that COMPANY's Revenue for the three months ending on December 31, 2022, is \$ 12,345.00 million.}.
These values are expressed in millions and rounded to two decimal places using conventional rounding techniques, providing the most prevalent and precise representation in financial reports.
We manually identified pertinent keywords and extracted values while training several individuals to assemble this dataset, ensuring each data point was labeled by four people to minimize labeling errors. 

\begin{figure}[ht]  
    \centering  
    \includegraphics[width=0.5\textwidth]{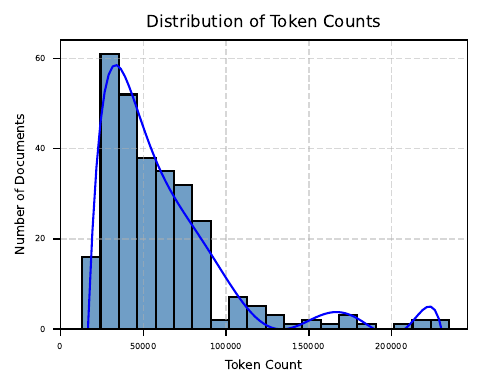}  
    \caption{Histogram of token counts in financial documents.}  
    \label{fig:token_distribution}  
\end{figure}

In selecting benchmark keywords, we prioritize their significance within financial reports.
We performed an intersection analysis on the essential keywords presented on two statistical websites publicly available from reputable organizations, MSN Money\footnote{\href{https://www.msn.com/en-us/money}{https://www.msn.com/en-us/ money}} and Google Finance\footnote{\href{https://www.google.com/finance/}{https://www.google.com/finance/}}, which showcase varying subsets of KPIs. 
We applied filtering criteria: keywords must exhibit ambiguity, be distributed throughout HLDs, and have values directly extractable from financial reports.
We identified a final set of 9 keywords (as presented in~\autoref{tab: 9keywords}) for further evaluation.
\autoref{fig:token_distribution} displays the token count distribution in FINE, with the largest document containing 234,900 tokens, the smallest document comprising 13,022 tokens, and an average of 59,464 tokens per document. 
\autoref{tab: revenue appearance} illustrates the specific representation of \textit{Revenue} in various companies' financial reports.
In FINE, we systematically document ambiguous expressions of all keywords across various companies.

\begin{figure}[htbp]
\includegraphics[width=\linewidth]{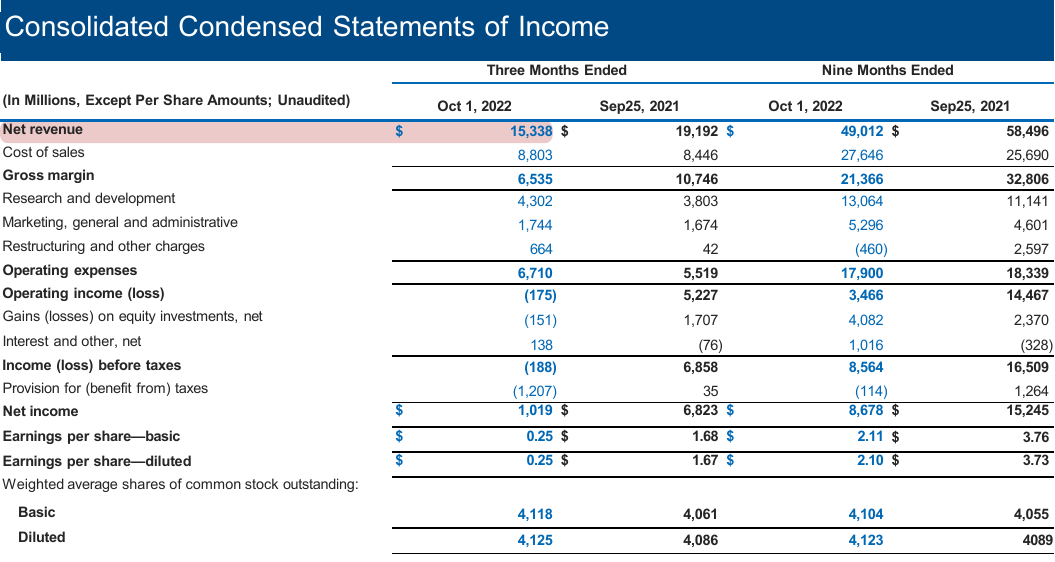}
\caption{Income statement of Intel in 2022-10-01 Quarterly report.}
\label{fig:Intel-10-Q-table}
\end{figure}

\begin{figure}[htbp]
\includegraphics[width=\linewidth]{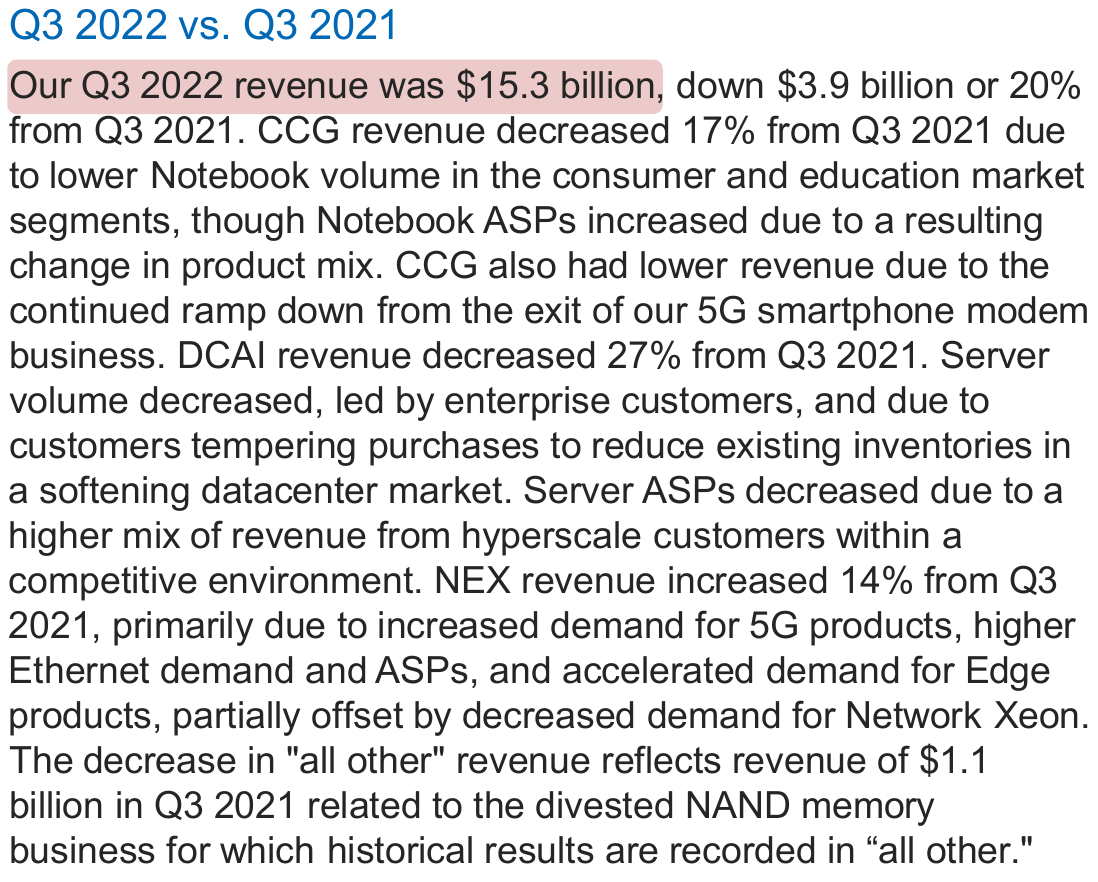}
\caption{A text description of Intel in 2022-10-01 Quarterly report.}
\label{fig:Intel-10-Q-text}
\end{figure}

\subsection{Wikipedia}
For this type of data, we chose the Wikireading-Recycled dataset~\cite{DBLP:journals/corr/abs-2011-03228}. 
This dataset is an improved version of the Wikireading dataset~\cite{DBLP:journals/corr/HewlettLJPFHKB16}, which includes a human-annotated test set. 
In this dataset, a Wikipedia page serves as the content, while the corresponding key and value are extracted from Wikidata. 
For example, from the \textit{Wikipedia of "In Search of Lost Time"} (Content), we can know that the \textit{main subject} (Key) of this novel is \textit{memory} (Value). 
From the human-annotated test set, we filtered out short samples with less than 10,000 tokens and those that would trigger safety restrictions in the text-davinci-003 model. 
After filtering, a total of 72 test samples remained for our evaluation.

For the Wikireading-Recycled dataset, the ground truth is in text form, and the predictions generated by LLMs often do not match the ground truth in terms of phrasing, despite conveying the same meaning. 
To evaluate the accuracy of LLM predictions, we combined the assessments of four human judgments and GPT-4's judgments. We then calculated the average of these evaluation results to determine the final metric.

\subsection{Scientific Papers}
For this type of data, we selected the MPP (Massive Paper Processing) dataset~\cite{polak2023flexible}.
In this dataset, scientific papers serve as the content, with chemical materials as the keys and their corresponding cooling rates as the values.
For example, from a paper "\textit{... the composition of $Al_{87}Ni_{9}Ce_{4}$ has the maximum cooling rate of nearly 1.02 × 10$4 K/s$...}" (Content), we can know that the \textit{cooling rate} (Key) of \textit{$Al_{87}Ni_{9}Ce_{4}$} is \textit{1.02 × 10$4 K/s$} (Value). 
We filtered out short papers and samples containing multiple values for the same key. Ultimately, 50 test samples remained for evaluation.

For the MPP dataset, the ground truth is numeric. This numeric value only appears in a unique form throughout the text. Therefore, we only needed to determine whether LLMs' predictions were consistent with the ground truth.

\section{Detailed Explanation of Four Challenges} \label{app: Three Challenges}

Hybrid documents, adeptly blending textual and tabular content, are widely used across diverse fields, and generally exhibit the following characteristics:

\textbf{1) Long Documents:}
The three types of HLDs selected for our experiments all encompass extensive text and tables related to relevant keywords, resulting in considerable length and complexity.
We found the average length of 59,464 tokens in experiments, equating to 14.5 times the max tokens of GPT-3.5 and 1.8 times that of GPT-4.

\textbf{2) Hybrid Content:} 
HLDs usually present the same information in different descriptive formats.
Take financial reports as an example, \autoref{fig:Intel-10-Q-table}  presents a table extracted from Intel's 2022Q3 financial report, while  \autoref{fig:Intel-10-Q-text} illustrates a textual description. Both representations convey the same keywords: \textit{Revenue} with different numerical precision.
To obtain keyword values accurately and effectively, it is essential to concurrently analyze both content types.

\textbf{3) Ambiguous Representation} and \textbf{4) Numerical Precision:} \textbf{In financial reports}, Varying expressions for the same keyword in HLDs across companies lead to ambiguity. \autoref{fig:Intel-10-Q-table} displays a table from Intel's 2022Q3 financial report, highlighting the \textit{Net Revenue} of \$15,338 million.
While, \autoref{fig:Intel-10-Q-text} features a textual description from the same report, indicating a \textit{Revenue} of \$15.3 billion. 
Both representations communicate the same KPI: \textit{Revenue}.
The presentation of identical information in varying formats is referred to as financial report ambiguity.  
\autoref{tab: revenue appearance} exemplifies this situation, demonstrating variations in \textit{Revenue} representation across different companies.
The textual and tabular contexts exhibit different numerical precision.
\textbf{In scientific papers}, cases are usually like this: the ground truth is ($Al_{0.07602085}Cu_{49.95655951}Gd_{0.01086012}Zr_{49.95655951}$,	10,	K/s), while the source in the original text is “\textit{... $(Cu_{50}Zr_{50})_{92}Al_{7}Gd_{1}$ is estimated to be 10Ks$^{-1}$ ...}”; ($Ni_{38}Zr_{62}$, 104, K/s), v.s. "\textit{... 104 K/s for the binary $Zr_{62}Ni_{38}$ ...}". And different numerical precision appears again.

These four structural properties render HLDs comprehension challenging for LLMs. To accurately obtain keywords and their values, both textual and tabular types must be analyzed concurrently for numerical precision.

\begin{table}[htbp]
    \centering
    \begin{tabular}{l|l}
        \hline

        \textbf{Category} & \textbf{Keywords} \\ \hline
        \multirow{4}{*}{\textbf{Income Statement}}
         & \textit{Revenue} \\
         & \textit{Operating Expense} \\
         & \textit{Net Income} \\
         & \textit{Earnings Per Share} \\ \hline
        \multirow{2}{*}{\textbf{Balance Sheet}}
         & \textit{Total Assets} \\
         & \textit{Total Equity} \\ \hline
        \multirow{3}{*}{\textbf{Cash Flow}}
         & \textit{Operating Activities} \\
         & \textit{Investing Activities} \\
         & \textit{Financing Activities}  \\ \hline
    \end{tabular}
    \caption{Nine Keywords in FINE.}
    \label{tab: 9keywords}
\end{table}

\begin{table*}[htbp]
    \centering
    \begin{tabular}{l|l}
        \hline
        \textbf{Corporation} & \textbf{\textit{Revenue}} \\ \hline
        Amazon & \textit{Total Net Sales}; \textit{Net Sales [span] Consolidated}; \textit{Consolidated [span] Net Sales}; \\
        AMD & \textit{Total net revenue}; \textit{Net revenue}; \textit{Total sales to external customers}; \\ 
        Apple & \textit{Total Net Sales} \\ 
        Autoliv & \textit{Consolidated net sales}; \textit{Net Sales}; \textit{Total Net Sales} \\
        BOEING & \textit{Revenues}; \textit{Total revenues} \\
        Cisco & \textit{Total revenue}; \textit{Product revenue: [span] total}; \textit{Revenue: [span] total}\ \textit{Revenue} \\
        Coca Cola & \textit{Net operating revenues} \\
        Dell & \textit{Total net revenue}; \textit{Total consolidated net revenue}; \textit{Net revenue} \\
        ebay & \textit{Net revenues}; \textit{Total net revenues} \\ 
        Intel & \textit{Net revenue}; \textit{Total net revenue} \\ 
        Meta Platforms & \textit{Revenue}; \textit{Total revenue} \\
        Microsoft & \textit{Revenue}; \textit{Total revenue} \\
        Nike & \textit{Revenues}; \textit{TOTAL NIKE, INC. REVENUES}; \textit{Total revenue}; \textit{Revenue} \\ 
        Nvida & \textit{Total revenues} \\
        Oracle & \textit{Total net revenues} \\ 
        Starbucks & \textit{Total revenue} \\ 
        State Street & \textit{Total revenues} \\
        Walmart & \textit{Total net revenues}; \textit{Total revenue} \\ \hline
    \end{tabular}
    \caption{The appearance of \textit{Revenue} in various company financial reports. We record the different occurrences of the selected keywords in FINE. [span] means that there are merged cells and indented forms in the table.}
    \label{tab: revenue appearance}
\end{table*}

\section{Detailed Experiment Settings}
\label{sc:exp_setting}
\textbf{Token Allocation:} We allocate tokens to accommodate the model's maximum sequence length and the requirements of each AIE module. 
The token allocations are presented in the \autoref{tab:token_allocation}.

\begin{table}[htbp]  
\centering  
\begin{tabular}{l|ll}  
\hline  
\textbf{Alloc.} & \textbf{\# Token} \\ \hline  
Max Seq. Length & 4,097\\  
Doc. Elem. & $\leq$ 2,000  \\ 
Doc. Seg. & $\leq$ 2,500  \\ 
Keyword & $\leq$ 50  \\ 
Summary & $\leq$ 500 \\ \hline  
\end{tabular}  
\caption{Token allocation}  
\label{tab:token_allocation}  
\end{table}   

\textbf{Prompts:} In AIE, there are many different types of prompts serving various AIE modules: question prompts, refine prompts, extraction prompts, and so on.
In these prompts, we apply our three prompt engineering.
\autoref{sec: Prompts} shows the details of the prompts.

\textbf{Retrieval Settings:} We select the top three document segments with the highest similarity scores for GPT-3.5.

\textbf{Embedding Model:} We use the sentence-transformers/all-mpnet-base-v2\footnote{\href{https://www.sbert.net/docs/pretrained_models.html}{https://www.sbert.net/docs/pretrained\_models.html}} model for computing embeddings. 
This model can handle a sequence length of 384 tokens. 

\begin{table}[ht]        
\centering        
\fontsize{8}{10}\selectfont 
\setlength{\tabcolsep}{2pt}
\begin{tabular}{l|cccc|c}        
\hline        
& \textbf{RETA 1\%} & \textbf{RETA 3\%} & \textbf{RETA 5\%} & \textbf{RETA 10\%} & \textbf{Average} \\ \hline    
\textbf{BOTH} & \textbf{0.6389} & \textbf{0.6938} & \textbf{0.7194} & \textbf{0.7451} & \textbf{0.6993} \\        
\textbf{TBL} & 0.5361 &	0.6014 &	0.6215 &	0.6465 &	0.6014 \\ \hline     
\end{tabular}  
\caption{Accuracy comparison between using both tabular and textual data (\textbf{BOTH}), and using only tabular data (\textbf{TBL}).}  
\label{tab:nec_both_data}
\end{table}

\begin{table}[ht]        
\centering        
\fontsize{8}{10}\selectfont 
\setlength{\tabcolsep}{2pt}
\begin{tabular}{l|cccc|c}        
\hline        
GPT-3.5 & \textbf{RETA 1\%} & \textbf{RETA 3\%} & \textbf{RETA 5\%} & \textbf{RETA 10\%} & \textbf{Average} \\ \hline    
\textbf{2019} & 0.6200&	0.6829&	0.7029&	0.7171&	0.6807 \\        
\textbf{2022} & 0.6417&	0.6917&	0.7222&	0.7361&	0.6979 \\ \hline     
\end{tabular}  
\caption{Accuracy comparison between samples from 2019 and 2022 using GPT-3.5.}  
\label{tab:cmp_gpt3_time}
\end{table}

\begin{table}[ht]        
\centering        
\fontsize{8}{10}\selectfont 
\setlength{\tabcolsep}{2pt}
\begin{tabular}{l|cccc|c}        
\hline        
GPT-4 & \textbf{RETA 1\%} & \textbf{RETA 3\%} & \textbf{RETA 5\%} & \textbf{RETA 10\%} & \textbf{Average} \\ \hline    
\textbf{2019} & 0.8543&	0.8857&	0.8914&	0.8914&	0.8807 \\        
\textbf{2022} & 0.7972&	0.8444&	0.8583&	0.8778&	0.8444 \\ \hline     
\end{tabular}  
\caption{Accuracy comparison between samples from 2019 and 2022 using GPT-4.}  
\label{tab:cmp_gpt4_time}
\end{table}

\section{Prompts}
\label{sec: Prompts}
  
\subsection{Summarization Prompts - Refine}
\label{app: Summarization Prompts - Refine}
The Refine strategy consists of two prompts: the Question Prompt and the Refine Prompt. These prompts are designed to guide LLMs in extracting and summarizing key information related to the given keywords from many segments.

\textbf{Question Prompt:} This prompt is designed to instruct the LLMs to generate an initial summary containing information related to the given keywords from the provided document segment. The content of the question prompt is as follows:

\begin{mdframed}
\begin{codeframe}
\begin{lstlisting} 

>>>>>Your Task:
Given a segment of a financial report and keywords.
You need to summarize the information related to the keywords.
All values must be in millions and round it to three decimal places using rounding rules.
>>>>>Example:
Financial report's segment: For company A in 2022Q3, the revenue is $1.2345 billion; the net income is $50.1245 million
-----
Keywords: Net income and revenue of company A in 2022Q3.
-----
Summary: For company A in 2022Q3, net income is $50.125 million, and revenue is $1,234.500 million.
>>>>>Question:
Financial report's segment: {document_segment}
-----
Keywords: {keywords}
-----
Summary: 
\end{lstlisting}
\end{codeframe}
\end{mdframed}

\textbf{Refine Prompt:} The refine prompt is designed to instruct LLMs to update the old summary by incorporating information related to the keywords from the newly provided document segment. The content of the refine prompt is as follows:

\begin{mdframed}
\begin{codeframe}
\begin{lstlisting} 

>>>>>Your Task:
Given a segment of a financial report, a summary of the previous segments and keywords.
You should combine the information related to the keywords to generate a new summary.
All values must be in millions and round it to three decimal places using rounding rules.
>>>>>Example:
Financial report's segment: For company A in 2022Q4, the net income is $5 billion.
-----
Old summary: For company A, the net income in 2022Q1 is $3.125 million; the net income in 2022Q2 is $123,123.000 million; the net income in 2022Q3 is $0.123 million.
-----
Keywords: Net income of company A in 2022.
-----
New summary: For company A, the net income in 2022 is $128,126.248 million.
>>>>>Question:
Financial report's segment: {document_segment}
-----
Old summary: {old_summary}
-----
Keywords: {keywords}
-----
New summary:
\end{lstlisting}
\end{codeframe}
\end{mdframed}

\subsection{Summarization Prompts - Map-Reduce}
\label{app: Summarization Prompts - Map_Reduce}  
The Map-Reduce strategy also consists of two prompts: the Map Prompt and the Reduce Prompt.

\textbf{Map Prompt}: This prompt is designed to instruct LLMs to generate a summary containing information related to the given keywords from the provided document segment. The content of the Map prompt is as follows:
\begin{mdframed}
\begin{codeframe}
\begin{lstlisting} 

>>>>>Your Task:
Given a segment of a financial report and keywords.
You need to summarize the information related to the keywords.
All values must be in millions and round it to three decimal places using rounding rules.
>>>>>Example:
Financial report's segment: For company A in 2022Q3, the revenue is $1.2345 billion; the net income is $50.1245 million.
-----
Keywords: Net income and revenue of company A in 2022Q3.
-----
Summary: For company A in 2022Q3, net income is $50.125 million, and revenue is $1,234.500 million.
>>>>>Question:
Financial report's segment: {document_segment}
-----
Keywords: {keywords}
-----
Summary:
\end{lstlisting}
\end{codeframe}
\end{mdframed}

\textbf{Reduce Prompt}: The Reduce prompt is designed to instruct LLMs to consolidate the summaries obtained from the Map process. The ``text'' in the prompt represents all the summaries generated by the Map process.
\begin{mdframed}
\begin{codeframe}
\begin{lstlisting} 

>>>>>Your Task:
Find the values of keywords in the given content.
If you can't find the value, please output "None". 
If you find the corresponding value, please express it in millions and round it to two decimal places using rounding rules.
>>>>>Example 1:
Content: For company ABC, total net sales for the three months ended June 25, 2022, were $65.135 billion.
-----
Keywords: Total net sales of ABC for the three months ended June 25, 2022.
-----
Result: 65,135.00
>>>>>Example 2:
Content: For company XYZ, total assets for the three months ended 2022.10.15 were $2.126 million.
-----
Keywords: Total assets of XYZ for the three months ended October 15, 2022.
-----
Result: 2.13
>>>>> Question
Content: {text}
-----
Keywords: {keywords}
-----
Result:
\end{lstlisting}
\end{codeframe}
\end{mdframed}

\subsection{Extraction Prompt}
\label{app: Extraction Prompt}  
\textbf{Extraction Prompt for GPT-3.5:} This prompt is designed to extract the numerical values corresponding to the specified keywords from the given content. If the value is not found, the prompt directs LLMs to output "None". If the value is found, it should be expressed in millions and rounded to two decimal places using rounding rules.
\begin{mdframed}
\begin{codeframe}
\begin{lstlisting}

>>>>> Your task:
Find the values of keywords in the given content.
If you can't find the value, please output "None". 
If you find the corresponding value, 
please express it in millions and round it to two decimal places using rounding rules.
>>>>> Example 1:
Content: For company ABC, Total Net Sales for the three months ended June 25, 2022, were $65.135 billion.
Keywords: Total Net Sales of ABC for the three months ended June 25, 2022.
Result: 65,135.00
>>>>> Example 2:
Content: For company XYZ, Total Assets for the three months ended 2022.10.15 were $2.126 million.
Keywords: Total Assets of XYZ for the three months ended October 15, 2022.
Result: 2.13
>>>>> Question:
Content: {text}
Keywords: {key_words}
Result:
\end{lstlisting}
\end{codeframe}
\end{mdframed}

\subsection{Numerical Precision Enhancement Prompts}
\label{app: Numerical Precision Enhancement Prompts}
The Numerical Precision Enhancement Prompts aim to improve the precision of extracted numerical values by guiding the LLMs to preserve the required level of precision. These prompts come in different variations, each adding or modifying specific aspects to achieve the desired precision:

\textbf{Naive}: This version of the prompt contains only a task description and task information. It does not provide explicit guidance on numerical precision.
\begin{mdframed}
\begin{codeframe}
\begin{lstlisting}

>>>>Your Task: 
Given a segment of a financial report and keywords. 
You need to summarize the information related to the keywords.
>>>>Question:
Financial report's segment: {document_segment}
-----
Keywords: {keywords}
-----
Summary: 
\end{lstlisting}
\end{codeframe}
\end{mdframed}

\textbf{Direct}: This version adds a precision requirement to the task description in the Naive prompt. It explicitly states that all values must be in millions and rounded to three decimal places using rounding rules.
\begin{mdframed}
\begin{codeframe}
\begin{lstlisting}

>>>>Your Task: ... All values must be in millions and round it to three decimal places using rounding rules ...
>>>>Question: ...
\end{lstlisting}
\end{codeframe}
\end{mdframed}

\textbf{Naive \& Shot}: Building on the Naive version, this prompt includes an input-output example. However, in this example, all the values are represented by variables x, y, and z. Therefore, this example doesn't provide any information about precision.
\begin{mdframed}
\begin{codeframe}
\begin{lstlisting}

>>>>Your Task: ...
>>>>Example:
Financial report's segment: For company A in 2022Q3, the revenue is $x billion; the net income is $y million.
-----
Keywords: Net income and revenue of company A in 2022Q3.
-----
Summary: For company A in 2022Q3, net income is $x million, and revenue is $y million.
>>>>Question: ...
\end{lstlisting}
\end{codeframe}
\end{mdframed}

\textbf{Direct \& Shot}: Combining the precision requirements from the Direct version and the example from the Naive \& Shot version, this prompt provides both explicit precision guidance and an example of the task, but without specific numerical values.
\begin{mdframed}
\begin{codeframe}
\begin{lstlisting}

>>>>Your Task: ... All values must be in millions and round it to three decimal places using rounding rules ...
>>>>Example:
Financial report's segment: For company A in 2022Q3, the revenue is $x billion; the net income is $y million.
-----
Keywords: Net income and revenue of company A in 2022Q3.
-----
Summary: For company A in 2022Q3, net income is $x million, and revenue is $y million.
>>>>Question: ...
\end{lstlisting}
\end{codeframe}
\end{mdframed}

\textbf{Naive \& Shot-Precision}: Building on the Naive \& Shot version, this prompt demonstrates how to preserve the required precision by using an input-output example with numbers.
\begin{mdframed}
\begin{codeframe}
\begin{lstlisting}

>>>>Your Task: ...
>>>>Example:
Financial report's segment: For company A in 2022Q3, the revenue is $1.2345 billion; the net income is $50.1245 million.
-----
Keywords: Net income and revenue of company A in 2022Q3.
-----
Summary: For company A in 2022Q3, net income is $50.125 million, and revenue is $1,234.500 million.
>>>>Question: ...
\end{lstlisting}
\end{codeframe}
\end{mdframed}

\textbf{Direct \& Shot-Precision}: This is the optimal prompt. It includes a precision requirement in the task description and an example demonstrating how to preserve the precision.
\begin{mdframed}
\begin{codeframe}
\begin{lstlisting}

>>>>Your Task: ...All values must be in millions and round it to three decimal places using rounding rules.
>>>>Example:
Financial report's segment: For company A in 2022Q3, the revenue is $1.2345 billion; the net income is $50.1245 million.
-----
Keywords: Net income and revenue of company A in 2022Q3.
-----
Summary: For company A in 2022Q3, net income is $50.125 million, and revenue is $1,234.500 million.
>>>>Question: ...
\end{lstlisting}
\end{codeframe}
\end{mdframed}

\begin{table*}
\centering  
\begin{tabular}{l|cccc|c}  
\hline  
\textbf{Naive} & \textbf{RETA 1\%} & \textbf{RETA 3\%} & \textbf{RETA 5\%} & \textbf{RETA 10\%} & \textbf{average} \\ \hline
Revenue        & 0.3056            & 0.3333            & 0.3438            & 0.3611             &                  \\  
Total Net Sales & 0.2361           & 0.2465            & 0.2604            & 0.2847             &                  \\ \hline  
\textbf{RPD}            & 25.64\%           & 29.94\%           & 27.59\%           & 23.66\%            & 26.71\%         \\ \hline
Total Equity   & 0.0260            & 0.0303            & 0.0390            & 0.0519             &            \\  
Total Stockholders' Equity & 0.0521 & 0.0556            & 0.0660            & 0.0799             &            \\ \hline  
\textbf{RPD}            & 66.90\%           & 58.82\%           & 51.48\%           & 42.35\%            & 54.89\%          \\ \hline
\end{tabular}  
\caption{Experimental results for the naive method in handling keyword ambiguity at different RETA levels}  
\label{tab:naive_results}  
\end{table*}

\begin{table*}
\centering  
\begin{tabular}{l|cccc|c}  
\hline  
\textbf{AIE} & \textbf{RETA 1\%} & \textbf{RETA 3\%} & \textbf{RETA 5\%} & \textbf{RETA 10\%} & \textbf{average} \\ \hline 
Revenue      & 0.8576            & 0.8611            & 0.8681            & 0.8889             &                  \\  
Total Net Sales & 0.8090         & 0.8299            & 0.8403            & 0.8542             &                  \\ \hline  
\textbf{RPD}          & 5.83\%            & 3.70\%            & 3.25\%            & 3.98\%             & 4.19\%           \\ \hline 
Total Equity & 0.4688            & 0.4861            & 0.5278            & 0.5556             &                  \\  
Total Stockholders' Equity & 0.5660 & 0.5938            & 0.6042            & 0.6493             &                  \\ \hline  
\textbf{RPD}          & 18.79\%           & 19.94\%           & 13.50\%           & 15.56\%            & 16.95\%          \\ \hline 
\end{tabular}  
\caption{Experimental results for AIE in handling keyword ambiguity at different RETA levels}  
\label{tab:our_results}  
\end{table*}

\section{Detailed Results of Keyword Ambiguity Experiment}  
\label{sec: Detailed Results of Keyword Ambiguity Experiment}
In this section, we present the detailed experimental results for both the naive method and AIE in handling keyword ambiguity. The results are shown for different RETA levels, as well as the average RPD for each comparison.  
  
\autoref{tab:naive_results} shows the experimental results for the naive method at different RETA levels. The results include comparisons between Revenue and Total Net Sales, as well as Total equity and Total stockholders' equity.  
\autoref{tab:our_results} displays the experimental results for AIE at different RETA levels. Similar to the naive method results, it includes comparisons between Revenue and Total Net Sales, as well as Total equity and Total stockholders' equity.

\section{Effect of Pre-training Data}
\label{sc:pre-train}
There is a common concern regarding LLMs: whether LLMs simply memorize the pre-training data, rather than possessing understanding and reasoning abilities. This concern raises the question of \textbf{whether pre-training data might interfere with the experimental results}.

The short answer is NO. We use the same pre-training model (e.g., GPT-3.5 or GPT-4) for each comparison, the result will not be affected by the pre-training data. And to know the impact of pre-training data containing documents on the results, we conducted relevant experiments in our study.

According to the available information, the datasets used for pre-training GPT3.5 and GPT4 were updated until September 2021. Therefore, we compared the 2019 and 2022 data in the FINE dataset. As shown in the \autoref{tab:cmp_gpt3_time} and \autoref{tab:cmp_gpt4_time}, the 2022 Average RETA score is higher than the 2019 score for GPT3.5. However, for GPT4, the 2019 Average RETA score is higher than in 2022. In both sets of experiments, the differences in Average RETA scores are not substantial. Therefore, we believe that the influence of pre-training data can be neglected for our experiments.

\section{Analysis of Computational Costs}
\label{sc:ana_cost}
For the analysis of time costs, we have already analyzed in \autoref{sc:ana strategy}.
For the analysis regarding the number of LLMs calls, it is related to the number of retrieved segments ($N_{seg}$), the maximum length of the document segment summary ($L_{sum}$). For the Refine strategy, the number of calls equals the number of retrieved segments plus one: $N_{call} = N_{seg} + 1$, which is four calls of GPT-3.5 for one financial report. For the Map-Reduce strategy, $N_{sum}$ represents the number of segment summaries, and $N_{mer}$ represents the number of LLMs calls required to merge segment summaries, $L_{mer}$ represents the length of summary that can be merged in one operation. In our experiment, only one merge operation is needed to merge all the segment summaries, so:$N_{sum} = N_{seg}$, $N_{mer} = 1$, $N_{call} = N_{seg} + 2$, which is five calls of GPT-3.5 for one financial report.

\section{Necessity of Considering Both Tabular Data and Textual Data}
\label{sc:nec_both_data}
In HLDs, there is many of information contained in tables, so there is a concern why not just using tabular data.
To evaluate the necessity of considering both tabular data and textual data.
We conducted experiments on FINE when using both tabular and textual data v.s. using only tabular data. 
The results are shown in the \autoref{tab:nec_both_data}. It indicates the necessity of using both modalities.

\end{document}